# EchoGuard: An Agentic Framework with Knowledge-Graph Memory for Detecting Manipulative Communication in Longitudinal Dialogue

**Ratna Kandala[1]**
**University of Kansas**
ratnanirupama@gmail.com

**Niva Manchanda[2]**
**University of Kansas**
n038k926@ku.edu

**Akshata Kishore Moharir[3]**
**University of Maryland**
akshatankishore5@gmail.com

**Ananth Kandala[4]**
**University of Florida**
ananthkandala46@gmail.com

## Abstract

Manipulative communication, such as gaslighting, guilt-tripping, and emotional coercion, is often difficult for individuals to recognize. Existing agentic AI systems lack the structured, longitudinal memory to track these subtle, context-dependent tactics, often failing due to limited context windows and catastrophic forgetting. We introduce EchoGuard, an agentic AI framework that addresses this gap by using a Knowledge Graph (KG) as the agent's core episodic and semantic memory. EchoGuard employs a structured Log-Analyze-Reflect loop: (1) users log interactions, which the agent structures as nodes and edges in a personal, episodic KG (capturing events, emotions, and speakers); (2) the system executes complex graph queries to detect six psychologically-grounded manipulation patterns (stored as a semantic KG); and (3) an LLM generates targeted Socratic prompts grounded by the subgraph of detected patterns, guiding users toward self-discovery. This framework demonstrates how the interplay between agentic architectures and Knowledge Graphs can empower individuals in recognizing manipulative communication while maintaining personal autonomy and safety. We present the theoretical foundation, framework design, a comprehensive evaluation strategy, and a vision to validate this approach.

## 1 Introduction

Manipulative communication tactics including gaslighting, guilt-tripping, and emotional coercion are pervasive in interpersonal interactions yet often difficult for individuals to recognize in real-time [Buss et al., 1987]. Awareness gaps, measurable discrepancies between an individual's emotional response to an interaction and their conscious recognition of manipulative intent persist even among individuals with high cognitive or emotional empathy [Thompson et al., 2022, Austin et al., 2007]. These gaps arise because manipulative language can operate through implicit mechanisms: the same phrase (e.g., "I'm just worried about you") can function as genuine support or as coercive control depending on relational context, power dynamics, and interaction history [Zhang et al., 2025, Cong, 2024].

While AI has advanced in sentiment analysis [Wang et al., 2024] and toxic language detection [Vidgen et al., 2021], existing systems face limitations when applied to interpersonal manipulation detection. First, context-dependency: current toxicity classifiers operate on utterance-level features and fail to capture how meaning emerges from relational dynamics. Second, implicit harm: manipulation often occurs through subtle linguistic patterns: presuppositions, conversational implicatures, and emotional appeals that are grammatically benign but pragmatically coercive. Third, subjective thresholds: what constitutes manipulation varies across individuals and relationships, requiring personalized rather than universal detection criteria.

Recent work has begun addressing these challenges. For instance, the MentalManip dataset [Wang et al., 2024] provides annotations for manipulation tactics in conversations, while LLM-based approaches [Khanna et al., 2025] explore introspective reasoning for multi-turn manipulation detection. However, these systems focus on automatic classification rather than scaffolding user awareness - a critical distinction when the goal is empowering individuals to recognize manipulative patterns themselves rather than outsourcing judgment to AI.

Furthermore, they fail to address the core memory challenge of agentic systems. Detecting longitudinal patterns like *intermittent reinforcement* or *escalating gaslighting* requires an agent to maintain a long-term, structured episodic memory of interpersonal history. Relying on an LLM's expanding context window is computationally inefficient, costly, and prone to catastrophic forgetting.

This is the precise gap addressed by the interplay of Knowledge Graphs and agentic systems—a core theme of this workshop. KGs provide a stable, structured, and queryable foundation for an agent's semantic memory (facts about manipulation) and episodic memory (the user's interaction history).

To address this, we introduce EchoGuard, a novel agentic AI framework designed to scaffold individual awareness of manipulative communication patterns. Unlike content moderation systems that flag overtly toxic language, EchoGuard operates as a "reflective analyzer." It employs an agentic loop where users log concerning interactions via a structured questionnaire that captures both the communication content and the user's emotional response. This questionnaire transforms a subjective experience into structured, machine-readable data. The agent transforms this subjective experience into a structured episodic Knowledge Graph. The agent then queries this KG against a semantic KG of manipulation tactics to identify specific, computationally-defined patterns. Finally, it generates a targeted, socratic prompt using a Large Language Model (LLM) to guide users toward self-discovery of the manipulative tactics present in their interaction.

The primary contributions of this work are threefold:

1. A novel agentic framework (Log-Analyze-Reflect) that leverages a Knowledge Graph as an external episodic memory to model and detect computationally tractable "Awareness Gaps."

2. A hybrid detection method combining graph-based reasoning (querying the episodic KG) with semantic analysis against a semantic KG of psychological constructs to identify six categories of manipulative tactics.

3. The design and implementation of the EchoGuard prototype, demonstrating the feasibility of using KG-grounded, constrained LLM interventions for sensitive personal safety applications.

## 2 The Proposed Framework: EchoGuard

We propose the EchoGuard agentic architecture to provide the cognitive scaffolding necessary to bridge the Awareness Gap which arises because manipulation exploits cognitive biases and emotional vulnerabilities, obscuring tactics over time. EchoGuard's core is a stateful agent that maintains accumulated context, materialized as a Knowledge Graph. This KG acts as a persistent working memory that victims often lack when cognitively overloaded. This agent autonomously orchestrates specialized analysis tools and selects adaptive intervention strategies to surface harmful patterns (Figure 1). By operationalizing this psychological insight, EchoGuard provides the framework needed to help users move past generalized distress and identify the specific "why" behind their harm.

### 2.1 Agent Architecture and Learning Mechanisms

Our framework's theoretical foundation, moving from felt experience to conscious recognition, is operationalized by the synergy between a stateful ReAct agent and a Knowledge Graph [Yao et al., 2023]. The KG serves as the agent's externalized, persistent memory, directly addressing the limited context window and catastrophic forgetting problems. The agent deploys four specialized tools: (1) a Structured Logger, which functions as an Agentic KG Enrichment tool, creating the cognitive space for reflection by populating the graph. To build the "mental schema" needed for "Pattern Recognition," it uses a (2) Pattern Detection Engine, which executes multi-hop queries against the KG. The agent's state is maintained by a (3) Context Analyzer, tracking user history via graph traversal. Crucially, to avoid psychological reactance and promote "Self-Discovery," a (4) Prompt Generator creates constrained, reflective LLM interventions grounded by the retrieved subgraphs that preserve user agency rather than issuing prescriptive directives.

**Agent Architecture and State Management:** Our agent's architecture is designed to operationalize the detection of complex psychological tactics. Manipulation patterns like gaslighting, moving goalposts, or love bombing/intermittent reinforcement are not static events; they are longitudinal strategies whose effects, such as learned helplessness or trauma bonding, accumulate over time. To capture this, the EchoGuard agent's persistent state $S$ is embodied in a Knowledge Graph, $G = (V, E)$. $V$ includes nodes for {`User, Other_Person, Interaction_Event, Emotion, Tactic, Phrase`}, and $E$ represents relationships like {`participated_in, felt_emotion, used_tactic, contains_phrase`}.

This graph state enables three critical capabilities: First, Contextual Adaptation uses the log history *by performing graph traversal* to identify these very longitudinal patterns, like intermittent reinforcement or escalations that a stateless analysis would miss. Second, Threshold Calibration adapts detection sensitivity based on feedback, refining its. understanding of the user's specific context. Finally, this state drives Tool Orchestration. For a new user, the agent may only flag clear-cut Gaslighting ("you're imagining things"). For a returning user, it can leverage the rich graph context to deploy semantic analysis, identifying subtle Guilt Induction ("after all I've done for you") or Emotional Blackmail tactics that are only evident in relation to past interactions.

## 2.2 Phase 1 - Structured Interaction Logging & Episodic KG Enrichment

In Phase 1, the agent's Structured Logger tool operationalizes the phenomenology of manipulation by transforming a subjective emotional experience into nodes and edges for the episodic KG. This tool captures three key psychological dimensions.

First, it records the Emotional State (using Plutchik's wheel) and the Cognitive Assessment (e.g., self-doubt), a key indicator of gaslighting, creating `Emotion` nodes (e.g., `Emotion:Self-Doubt`) and `Cognition` nodes (e.g., `Cognition:Self-Doubt`), linking them to the core `Interaction_Event` node. Second, it logs the Communication Content; this logging itself acts as a therapeutic intervention, initiating "affect labeling" [Lieberman et al., 2007] and cognitive processing.

Crucially, the tool derives a computational signature for the Awareness Gap by querying the KG for `Interaction_Event` nodes with a high-distress `felt_emotion` edge but a low-confidence `articulated_cause` edge. This captures the felt-sense of manipulation that the user cannot yet consciously recognize.

## 2.3 Phase 2 - Reasoning over the KG for Pattern Detection

In Phase 2, the Pattern Detection Engine's primary function is psychological validation: it provides the external, legitimate recognition that victims of manipulation are often denied. To achieve this, the engine operationalizes six manipulation tactics (e.g., Gaslighting, Guilt Induction, Emotional Blackmail) using a hybrid methodology that fuses graph-based reasoning with semantic analysis.

First, a semantic KG defines the psychological markers for each tactic (e.g., Gaslighting $\xrightarrow{\text{indicated\_by}}$ High_Self_Doubt_Marker, Gaslighting $\xrightarrow{\text{indicated\_by}}$ Reality_Denial_Marker). Second, the agent executes a query across the user's episodic KG to find subgraphs matching these markers. For instance, it searches for {Interaction_Event $\xrightarrow{\text{felt\_emotion}}$ Emotion:Self-Doubt} *and* {Interaction_Event $\xrightarrow{\text{contains\_phrase\_similar\_to}}$ Reality_Denial_Bank}. This is supplemented by a semantic component using Sentence-BERT to match user-submitted phrases against curated, expert-validated pattern banks. The resulting weighted confidence scores are framed not as a technical verdict but as psychological signals that validate the user's experience, providing transparent evidence for why they feel distressed.

## 2.4 Phase 3 - KG-Grounded Reflective Prompts

In Phase 3, the agent moves from detection to intervention, operationalizing principles from motivational interviewing to scaffold user insight. The goal is to avoid psychological reactance; a directive like "This person is guilt-tripping you" removes agency, whereas a reflective question ("What assumptions about obligation did that phrase rely on?") invites curiosity and self-discovery. This principle is implemented by the Reflective Prompt Generator, which uses a three-layer constraint architecture to safely leverage LLMs.

- Template Constraints: The system prompt explicitly prohibits diagnostic language ("abuser") and directive advice ("you should leave"), directly aligning with the psychological goal of focusing on behaviors, not labels, to prevent defensive reactions.

- Grounding Strategy: The LLM is grounded with structured data retrieved from the KG query (the detected pattern, user emotions, and specific phrases *from the identified subgraph*). This solves the context window problem by providing only the most relevant, multi-step context, rather than a long, unstructured chat history.

- Adaptive Learning Loop: The agent's state is updated based on user engagement and feedback. This operationalizes the psychological

necessity of respecting individual differences. By adapting to a user's communication style (e.g., reducing sensitivity for a user who finds it unhelpful), the agent avoids imposing a rigid interpretive framework and models the very flexibility that manipulation lacks.

## 3 Proposed Evaluation Strategy

We propose a comprehensive evaluation methodology to validate the framework's efficacy and identify limitations.

### 3.1 Experimental Design

We will conduct a Multi-Arm Randomized Controlled Trial (RCT) to isolate component contributions, comparing four conditions: (1) the Full EchoGuard Log-Analyze-Reflect workflow, (2) Structured Logging Only, (3) a Psychoeducation Baseline (generic content), and (4) a Control reflection task. Stimuli will consist of validated conversational vignettes depicting manipulation across diverse contexts (e.g., workplace, romantic), alongside foil vignettes featuring assertive but non-manipulative communication to assess false positive rates. The Primary outcome is a Manipulation Recognition Task measuring participants' ability to identify tactics, select specific phrasal evidence, and assess impacts like emotional coercion. Secondary outcomes include metacognitive awareness, transfer of skills to novel patterns, and safety measures (e.g., monitoring for increased anxiety).

### 3.2 Baseline Comparisons

To demonstrate EchoGuard's unique contribution, we will establish three critical baseline comparisons. First, we will test its performance against standard toxic language detectors to quantify how often subtle manipulation is missed by systems not trained on these psychological constructs. Second, we will compare our structured, constrained-agent workflow against a zero-shot prompt analysis to evaluate the safety and efficacy benefits of our guided approach versus direct, unconstrained prompting. Third, we will compare our KG-based memory agent against an agent using a standard vector database or flat log file for memory, to quantify the performance, latency, and accuracy gains from the structured reasoning capabilities of the graph.

### 3.3 Ablation Studies

Further, ablation studies will be performed to precisely isolate the contribution of each architectural component. These studies will quantify the added value of graph-based episodic memory over free-form reflection, the performance gain from our hybrid KG/semantic detection engine versus simple keyword matching, and the efficacy of our KG-grounded prompt generation compared to generic psychoeducational content.

### 3.4 Failure Mode Analysis

A rigorous Failure Mode Analysis is planned to explicitly investigate model limitations. This includes assessing false positives (e.g., misidentifying healthy, assertive communication as manipulative) and analyzing the impact of cultural variation and context dependence on detection accuracy. We will also evaluate user experience failures, such as interventions that inadvertently increase distress, to refine the agent's safety constraints.

## 4 Ethical Considerations and Limitations

Safety Design Principles: The agent's architecture embeds several core Safety Design Principles. The system is fundamentally constrained to describe communication patterns, not label people (e.g., as "abusers"), a key psychological safeguard against defensiveness. Interventions are limited to reflective questions to foster self-discovery, explicitly prohibiting directive advice (e.g., "you should leave"). The framework is positioned as an educational, not diagnostic, tool, with prominent links to crisis resources and no storage of personally identifiable information.

Acknowledged Limitations: We acknowledge several key limitations. The initial evaluation relies on controlled vignettes, which lack the emotional investment and incomplete information of real-world interactions. While our KG memory mitigates this, it is only as good as the data the user logs. Furthermore, the agent lacks access to the full relationship history that provides essential context. The detection patterns, derived from Western psychological literature, require significant cultural specificity validation for the semantic KG. Finally, there is an inherent oversimplification risk in this approach, and the harm of false positives (incorrectly flagging healthy communication) remains a primary concern.

Commitment Against Misuse: We affirm our commitment against misuse. This framework is architected as a first-person, self-reporting tool, which inherently limits its applicability for surveillance. The research and resulting models will not be used to develop third-party monitoring or tracking applications.

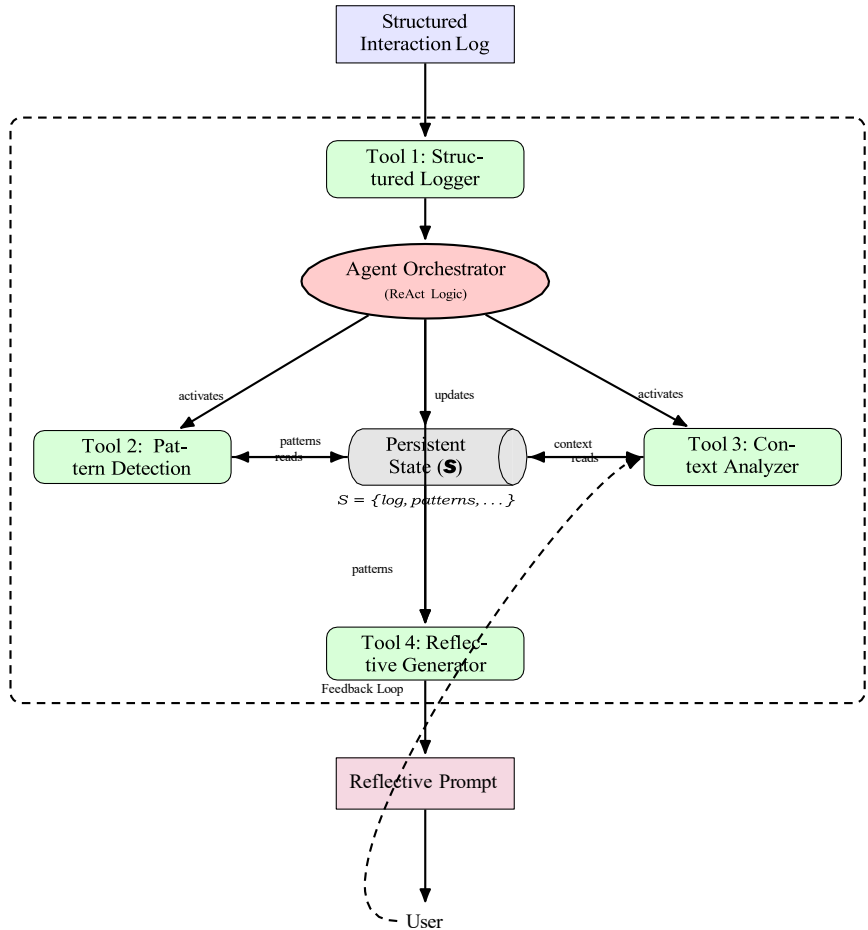

Figure 1: The EchoGuard Framework Architecture. The central agent orchestrates specialized tools, maintaining a persistent state (*S*) to analyze user logs. It uses a constrained prompt generator to create reflective interventions, which in turn update the agent's context via an adaptive learning loop.

## 5 Conclusion, Expected Impact and Future Directions

EchoGuard demonstrates a path for agentic AI in sensitive socio-emotional domains. Its core contributions are operationalizing psychological constructs within a Knowledge Graph memory and balancing automation with human agency through KG-grounded, constrained, reflective LLM interventions. Future work will focus on longitudinal, real-world deployment and rigorous cross-cultural validation of its semantic KG's detection patterns. This work establishes design principles for AI that scaffolds, rather than supplants, human judgment in complex social domains.

## A Appendix

### A.1 Definitions

**Interpersonal manipulation:** Communicative acts intended to influence another's beliefs, emotions, or behaviors through deception, coercion, or exploitation of psychological vulnerabilities, typically prioritizing the manipulator's interests while obscuring this intent ([Simon and Foley, 2011]).

**Cognitive Empathy (also called perspective-taking or empathic accuracy):** The ability to take the mental perspective of others, allowing one to make inferences about their mental or emotional states ([Shamay-Tsoory, 2011]). It involves understanding how a person feels and what they might be thinking without necessarily engaging with their emotions, a more rational and logical process ([Davis, 1983]).

**Emotional Empathy (also called affective empathy):** The ability to share the emotional experiences of others, a visceral reaction to their affective states ([Shamay-Tsoory, 2011]). Our ability to empathize emotionally is based on emotional contagion, being affected by another's emotional or arousal state. This is when you literally feel the other person's emotions alongside them, as if you had 'caught' the emotions.

**Agentic AI:** An AI system that autonomously executes multi-step reasoning processes, maintains state across interaction cycles, and adapts its behavior based on intermediate outputs without requiring human intervention at each step ([AWS, 2025]).